\documentclass[dvipsnames,format=sigconf]{acmart}
\usepackage{mathtools} 
\usepackage[ruled,linesnumbered,vlined]{algorithm2e}
\usepackage{subcaption} 
\usepackage{booktabs}
\usepackage{todonotes}

\renewcommand{\epsilon}{\varepsilon}

\newcommand{\sobol}{Sobol\kern-0.15em\'{ } }
\usepackage{siunitx}

\AtBeginDocument{%
  \providecommand\BibTeX{{%
    \normalfont B\kern-0.5em{\scshape i\kern-0.25em b}\kern-0.8em\TeX}}}

\setcopyright{acmcopyright}

\acmDOI{10.1145/nnnnnnn.nnnnnnn}
\acmISBN{978-x-xxxx-xxxx-x/YY/MM}
\acmConference[GECCO '24]{The Genetic and Evolutionary Computation Conference 2024}{July 15--19, 2024}{Melbourne, Australia}
\acmYear{2024}
\copyrightyear{2024}

\begin{document}

\title{Understanding fitness landscapes in morpho-evolution via local optima networks}

\author{Sarah L. Thomson}
\affiliation{%
  \institution{Edinburgh Napier University}
  \city{Edinburgh}
  \country{Scotland, UK}}
  \email{s.thomson4@napier.ac.uk}

\author{Léni Le Goff}
\affiliation{%
  \institution{Edinburgh Napier University}
  \city{Edinburgh}
  \country{Scotland, UK}}
  \email{l.legoff2@napier.ac.uk}

\author{Emma Hart}
\affiliation{%
  \institution{Edinburgh Napier University}
  \city{Edinburgh}
  \country{Scotland, UK}}
  \email{e.hart@napier.ac.uk}

\author{Edgar Buchanan}
\affiliation{%
  \institution{University of York}
  \city{York}
  \country{England, UK}}
  \email{edgar.buchanan@york.ac.uk}

\renewcommand{\shortauthors}{Thomson et al.}

\begin{abstract}
Morpho-Evolution  (ME) refers to the simultaneous optimisation of a robot's design \textit{and} controller to maximise performance given a task and environment. Many genetic encodings have been proposed which are capable of representing design and control.
Previous research has  provided empirical comparisons between encodings in terms of their performance with respect to an objective function and the diversity of designs that are evaluated, however there has been no attempt to \textit{explain} the observed findings. We address this by applying Local Optima Network (LON) analysis to investigate the structure of the fitness landscapes induced by three different encodings when evolving a robot for a locomotion task, shedding  new light on the ease by which different fitness landscapes can be traversed by a search process. This is the first time LON analysis has been applied in the field of ME despite its popularity in combinatorial optimisation domains; the findings will facilitate design of new algorithms or operators that are customised to ME landscapes in the future.
\end{abstract}


\maketitle

\section{Introduction }

Ever since the pioneering work of Karl Sims almost 30 years ago \cite{sims1994evolving}, evolutionary approaches to simultaneously optimise both the designs (i.e. the body) and controllers of robots have gathered pace. We term this morpho-evolution (ME) in this article. A variety of approaches now exist ranging from those that apply evolution purely in simulation \cite{lipson2016difficulty}, through hybrid approaches that combine hardware and software evolution \cite{hale2019robot} to those that evolve purely in hardware \cite{brodbeck2015morphological}. Evolutionary frameworks that facilitate ME must consider the challenging question of how to represent both designs and controllers in a form that can be evolved. Direct encodings such as directed trees \cite{veenstra2020different} provide a one-to-one mapping between a genotype encoding the design and control that is realised as the phenotype, and have been postulated to facilitate the fine-tuning of parameters \cite{veenstra2022effects}. 
On the other hand, indirect encodings such as Compositional Pattern Producing Networks (CPPNs) \cite{stanley2007compositional} and L-Systems \cite{lindenmayer1992grammars} result in many-to-one genotype-phenotype mappings which have several potential benefits, e.g. in being able to produce repeating patterns. This characteristic is generally useful in robotics as it can facilitate the creation of symmetrical designs, e.g. with equal numbers of actuators on each side of a body.
However, when using an indirect encoding, the genotype and the phenotype spaces are not necessarily isometric: small changes in the genotype can lead to major changes in the phenotype, and vice-versa \cite{hart2022artificial,angus2023practical}). Several authors have also noted that the evolutionary progress using an indirect encoding is slower than that of a direct one (e.g. \cite{le2020pros}). While there have been several studies that have compared direct and indirect encodings for jointly optimising design and control in terms of the level of performance reached (i.e. objective fitness) and the diversity of designs of evolved robots~\cite{veenstra2017evolution, miras2018search, de2020comparing, veenstra2020different, veenstra2022effects}, to the best of our knowledge there has been no attempt to \textit{explain} these empirical observations. That is, there has been no attempt to study the fitness landscapes induced by a choice of encoding to shed light on the ease by which these landscapes can be traversed during a search process. Understanding these landscapes could help explain empirical observations regarding performance or diversity, but could also facilitate the design of new algorithms or operators that are customised to the type of landscapes found.

The novel contribution of this paper is to apply \textit{Local Optima Network} (LON) analysis  \cite{ochoa2014local}
to analyse the structure of the fitness landscapes created by different encodings when jointly optimising the body and control of a simulated robot. LONs are a technique first introduced in 2008 \cite{ochoa2008study} which 
enables the information contained in a search space to be compressed into a single graph. Analysis of the graph enables the calculation of a set of metrics which characterise the structure of a landscape. The technique has proved very popular in the combinatorial optimisation community \cite{ochoa2018mapping,ochoa2019local,munoz2022local}, and more recently with continuous functions \cite{mitchell2023local} but has not been used in evolutionary robotics. We pose the question  \emph{"To what extent can applying LON analysis to fitness landscapes induced by different encodings in morpho-evolution provide an explanation for (1) observed performance and (2) the extent to which the search-space is explored".}
To answer this, we apply LON analysis to landscapes induced by three different encodings (a direct encoding producing a tree, L-System, CPPN) for an ME locomotion task, shedding new light on the effect of solution encoding on search-space \emph{navigability}. We define navigability as the ability for the search to a) consistently escape local optima, and b) explore a large number of genotypes. This is the first time that LON analysis has been applied within Evolutionary Robotics (ER). Furthermore, it is also novel in applying LON analysis to representations in which there is no direct equivalence between genotype and phenotype: there is only one other attempt at this that we are aware of in which LON analysis is applied to Linear Genetic Programming \cite{hu2023phenotype}. The results indicate that L-System indirect encoding leads to better fitness landscape navigability; this helps to explain its superior performance in terms of fitness in some empirical studies \cite{veenstra2020different,veenstra2022effects}. In contrast, the CPPN encoding is shown to induce lower navigability --- in fact, it seems to have landscapes which exhibit large numbers of low-quality local optima which are difficult to escape. 

\section{Related Work}

Recent years have seen a significant increase in efforts focused on the joint optimisation of both design and control of robots using evolutionary methods~\cite{auerbach2014robogen,brodbeck2015morphological,lipson2016difficulty,miras2020environmental,hale2019robot,li2023evaluation}. Much of this work is inspired by the seminal work of Pfiefer \& Bongard~\cite{pfeifer2006body} `How the Body Shapes the Way We Think' which argues that thought is not independent of the body but is tightly constrained, and at the same time enabled, by it --- suggesting the evolution of design and cannot control be considered independently. Several evolutionary frameworks have been proposed to simultaneously evolve design and control~\cite{miras2018search,hale2019robot, gupta2021embodied,miras2020environmental,li2023evaluation}. Regardless of the choice of framework, a decision must be made regarding how to encode both design and control on a single genome.
A number of encodings have been suggested, which can be categorised as \textit{direct} or \textit{indirect}. In the former, typical examples include the use of directed trees \cite{auerbach2014robogen,veenstra2017evolution,veenstra2020different} which for example specify how modules are connected to create a design, as well as the control parameters of each module. Indirect encodings require a mapping from genotype to phenotype. The use of CPPN \cite{stanley2007compositional,cheney2014unshackling,kriegman2020scalable} is common here: a particular advantage of this type of encoding is that it can produce repeating patterns, a characteristic which is often beneficial in producing symmetric robots which facilitate locomotion. An alternative indirect encoding is an L-System \cite{lindenmayer1992grammars} in which an evolved grammar produces rules that specify the robot \cite{miras2018effects,miras2019effects}. A number of studies have examined the effect of the choice of encoding on performance (i.e. objective fitness) in joint optimisation of design and control~\cite{veenstra2017evolution,veenstra2022effects}.
In \cite{veenstra2020different}, the effect of the encoding on solution diversity is also examined. Miras {\em et. al.}~\cite{miras2021constrained} compare two generative encodings in terms of how the choice of encoding influences the phenotype of evolved robots.  In \cite{pigozzi2023morphology}, the extent to which the encoding influences learning ability in joint optimisation. However, there have been very few attempts to relate observed empirical performance metrics to the underlying fitness landscape induced by the choice of encoding, particularly in morpho-evolution. Some early literature in fitness landscape analysis in ER focused solely on the controller of the robot \cite{dittrich1999dynamical}. In \cite{smith2002neutrality} a simple form of morpho-evolution was investigated in which the sensor placement of a fixed robot body is evolved along with the controller. The authors investigate ruggedness and neutrality within this landscape.  Naya-Varela {\em et. al.} \cite{naya2023guiding} propose an evolutionary algorithm for 
\textit{growth-based morphological development} in which the robot structure `grows' overtime following a fixed developmental schedule.  In this work, the authors employed a graph-based tool called Search Trajectory Networks (STN) \cite{ochoa2021search} which enables the trajectory of the best solutions found by a population-based algorithm to be visualised over time. The STN reveals information about how the algorithm traverses the underlying fitness landscape. However, this algorithm does not fall in the class of `morpho-evolution' algorithms considered in this article as the design changes according to a pre-determined schedule, rather than being evolved. Outside of ER, LONs have been used to bring new insights into the optimisation dynamics of a wide variety of different problems e.g. in combinatorial optimisation \cite{ochoa2008study,pavelski2021local}, neural architecture search (NAS) \cite{ochoa2022neural}, in algorithm-selection \cite{bozejko2018local} and in continuous landscapes \cite{ochoa2014local}.  All of the aforementioned examples apply LON analysis to encodings in which there is a one-to-one mapping between genotype and phenotype.  The only example of which we are aware in which LON analysis is applied to a scenario in which there is a many-to-one mapping is a recent analysis of Linear Genetic Programming \cite{hu2023phenotype}. In summary, this article adds to the field of ER by using LONs for the first time to understand fitness landscapes, in the context of jointly optimising design and control using different encodings.

\section{Preliminaries}
This section provides a brief primer on LONs and defines the terms that are used throughout the remainder of the article in this respect, given that this is the first time they have been used in the context of Evolutionary Robotics. LONs \cite{ochoa2008study} are a means to study the global structure of a fitness landscape. In a LON, nodes represent local optima in a fitness landscape, and edges between nodes are search transitions between them. By \emph{search transition}, we mean that two locations in the configuration space were linked during a metaheuristic search: in our study, a search transition between nodes is an application of perturbation followed by hill climbing. It is therefore a sequence of search operations applied to the source node which results in the destination node. LONs can be created by sampling local optima in a fitness landscape using Iterated Local Search (ILS) \cite{lourencco2003iterated}. We now define some key terminology:

\paragraph{Neighbourhood.} The \emph{neighbourhood} of a solution, $s_i$, are the solutions which are adjacent to $s_i$ according to a neighbourhood function: $N(s)$. In this work, the notion of adjacency depends on the design encoding type: the particulars of these are detailed in Section \ref{mutationoperators}. 

\paragraph{LON nodes.} A local optimum has superior or equal fitness to its neighbours according to a fitness function \emph{f}. In this work, we do not exhaustively search the neighbourhood: this would be computationally infeasible. Instead, we consider that a solution $lo_i$ is a local optimum if it has superior or equal fitness to its \emph{sampled} neighbourhood $SN$. Formally: \(\forall{ n \in SN(lo_i)}:\) $f(lo_i) >= f(n)$ (assuming maximisation, as is the case for this study) where $SN(lo_i)$ is the sampled neighbourhood, $n$ is a particular neighbour. The nodes in a LON, $LO$, are the local optima as just defined. 

\paragraph{LON edges.} There is an edge from local optimum $lo_i$ to local optimum~$lo_j$, if $lo_j$ can be obtained after applying a random perturbation to $lo_i$ followed by local search, and $f(lo_j) \geq f(lo_i)$. In LON terminology, these are called \emph{escape} edges \cite{verel2012local}. The edges are called {\em monotonic} because they record only non-deteriorating, directed connections between local optima. Edges are weighted with the frequency of transition: the number of times during searches that $lo_j$ was reached by applying perturbation then local search to $lo_i$. The set of edges is denoted by $E$.  

\paragraph{Local optima network (LON).} A local optima network, LON = $(LO,E)$, consists of nodes $lo_i \in LO$ which are the local optima, and edges $e_{ij} \in E$ between pairs of nodes $lo_i$ and $lo_j$ with weight $w_{ij}$ iff $w_{ij} > 0$. A LON which only includes neutral or improving transitions between local optima is a \emph{monotonic} LON, or MLON; we construct these for the present work. For the interested reader, detailed descriptions of MLONs can be found in previous literature~\cite{ochoa2017understanding}. 

\section{Methodology}

\subsection{Task and Environment}
\begin{figure}
\centering
  \fbox{\includegraphics[width=0.80\linewidth]{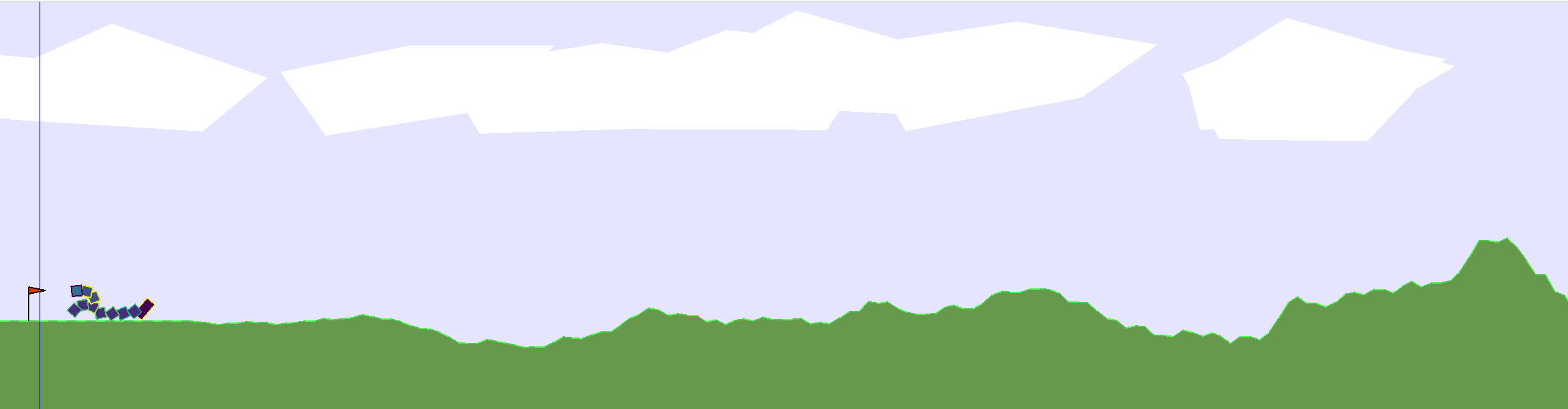}}
\caption{An example robot attempting to move horizontally through the virtual environment. Each shape is a module, and each module has its own controller.}
\label{fig:robot-env}
\end{figure}

We conduct LON analysis on a joint optimisation task taken from  Veenstra \emph{et al.} \cite{veenstra2022effects}
which focused on empirical performance comparisons of robots evolved using different encodings. The goal is to evolve a robot that needs to traverse a virtual landscape consisting of a set of gentle hills. The objective is to maximise the distance travelled. We use an environment called gym2D which facilitates robot design evolution for OpenAI Gym \cite{brockman2016openai}; its foundation was the bipedal walker environment \footnote{\url{https://github.com/openai/gym/blob/master/gym/envs/box2d/bipedal_walker.py}}. The beginning section of the environment we use is depicted in Figure~\ref{fig:robot-env}. A virtual robot begins at the left of a rectangular box containing a virtual 2D landscape and the objective is to move as far to the right of the box as possible; the horizontal axis for the box is associated with a scale between 0 and 100 (0 is no movement; 100 is reaching the end) and the fitness of a robot is how far along this axis it travels. There is also a "kill-switch" implemented to reduce the computational cost: if a robot travels slower than a minimum speed (set at 0.04) then the attempt terminates and the robot is ascribed a default fitness of 5.0. The same environment configuration is used across all the experiments. A robot can be seen attempting to travel across the landscape in Figure \ref{fig:robot-env}.

\subsection{Robots}

A \textit{robot} is defined by a set of \textit{modules} combined to produce a 2D~shape.
Modules are either circular or rectangular and each has parameters defining its size. The allowable size ranges for modules are as follows: a width and height between 0.5 and 1, and (in the case of circle shapes) a radius between 0.25 and 0.5. Modules also each have a controller, as described shortly. A list of eight modules is kept in memory and these are selected to be used as nodes to build the robot. For each individual run, a new initial \textit{module list} is generated of four circles and four rectangles. The radius of the circles is randomly generated between the bounds just mentioned; similarly, the width and height of the rectangles are randomised. The parameters of the modules are mutated during the evolutionary runs. Each virtual robot has one controller per module. A controller is encoded as a sine wave with parameters: amplitude (\(\alpha\), frequency \(\theta\), phase \(\delta\), and offset \(\epsilon\),

\begin{equation}
y(t) = \alpha sin(\theta t + \delta) + \epsilon
\end{equation}\label{eq:controller}

The ranges for the parameters are: \(\alpha\) and \(\delta\) can vary between -1 and +1; \(\theta\) between -0.1 and +0.1; and \(\epsilon\) between \(-\pi\) to \(+\pi\). These parameters are also subject to mutation (details of mutation are found later in Section \ref{mutationoperators}. Finally, the maximum depth of the tree which represents a robot is seven, while the maximum size (number of allowable modules) is 40. These parameters are taken directly from the literature \cite{veenstra2022effects}.

\subsection{Encodings}\label{encodings-description}
\begin{figure}[htb!]
\centering
  \fbox{\includegraphics[width=0.35\textwidth]{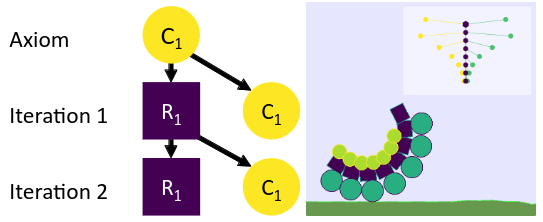}}
\caption{On the left: directed tree created by an L-System encoding after two iterations. The nodes labelled $C_1$ and $R_1$ show a circle and
rectangle module, respectively. On the right: an example
of a virtual 2D robot; its graph is visualised in the top
right. Figure is from \cite{veenstra2022effects}}
\label{fig:tree}
\end{figure}

The phenotype of every virtual creature in Gym2D is represented as a tree (see the left sub-plot of Figure \ref{fig:tree}). We evaluate three different encodings to produce this tree. The first encoding is direct, therefore exactly represents the tree. Two encodings are indirect therefore there is a mapping from genotype to phenotype. The same three encodings were used in \cite{veenstra2020different}.
For the indirect encoding, the tree starts from an axiom and then expand iteratively until reaching a maximum depth. Each node can have a maximum of three nodes attached. The following section describes how each encoding generates a tree.

\paragraph{Direct encoding.} The direct robot design encoding is a directed tree structure which represents the topology of the robot's design.  For each node that makes up a robot, the encoding describes: the node's index in the tree; the index of the module from the module list it represents; the index of its parent node; and the parent connection coordinates. 

\begin{figure}[htb!]
\centering
  \fbox{\includegraphics[width=0.45\textwidth]{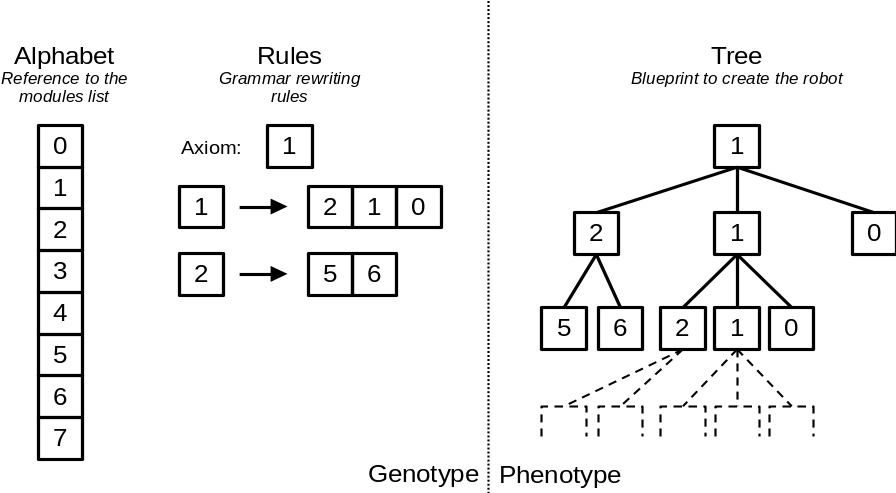}}
\caption{Diagram representing how an L-system works}
\label{fig:lsystem-dia}
\end{figure}

\paragraph{L-System.} For the L-System encoding, a parametric approach is used where symbols and grammar rules are used (see figure~\ref{fig:lsystem-dia}). A set of symbols is specified via an alphabet. The alphabet is composed of eight symbols, each of which corresponds to a module in the modules list. A set of rules, known as \emph{grammar rewrite rules}, are applied to the symbols in order to construct the tree. A rule is composed of an input symbol and substitution symbols, traditionally, the input symbols are replaced by the substitution symbols. In this system, starting from an axiom symbol, the rules are used to add new leaves (each leaf corresponds to one of the substitution symbols) to the tree from a previously created node (the input symbol). 

\begin{figure}[htb!]
\centering
  \fbox{\includegraphics[width=0.45\textwidth]{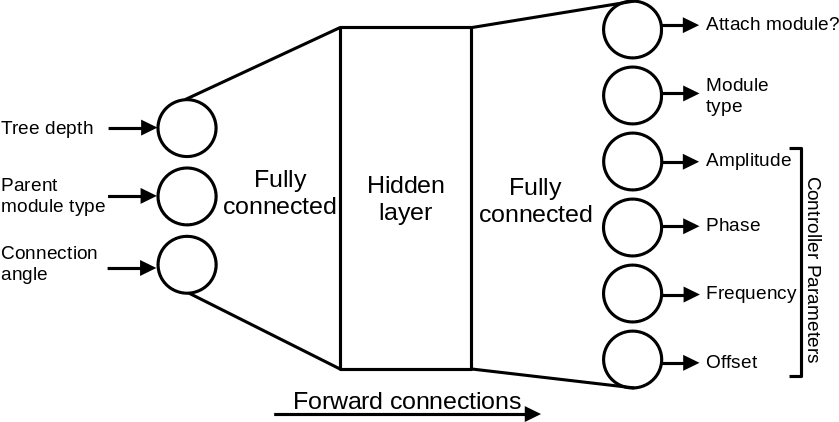}}
\caption{Diagram representing a CPPN network}
\label{fig:cppn-dia}
\end{figure}

\paragraph{CPPN} CPPNs are neural networks specifically designed to produce patterns where the inputs are spatial coordinates. Each network node is a function representing a basic pattern such as Gaussian for axial symmetry, sinusoidal for repetition, or sigmoid for central symmetry. As CPPN is similar to a feed-forward neural network it is classically trained using neuro-evolution. In this study, we make use of the Neuro-evolution through Augmenting Topologies (NEAT) \cite{stanley2002evolving}. To create the tree, the CPPN is queried incrementally for each connection site, i.e. where a node can be potentially attached. The CPPN takes three inputs: the depth of the connection site in the tree, the parent index in the module list and the angle of connection. Thus, the CPPN produces 6 outputs:  whether a module should be attached or not, the type of module, i.e. the index in the modules list, and four outputs corresponding to the controller parameters (see figure~\ref{fig:cppn-dia}).

\subsection{LON Construction}

Following the convention in the LON literature (see, for example: \cite{treimun2020modelling,veerapen2017modelling,thomson2023randomness,pavelski2021local}), we sample local optima  by executing runs of Iterated Local Search (ILS). ILS \cite{lourencco2003iterated} is a single-point meta-heuristic which conducts repeated cycles of applying large perturbations followed by local search. Although algorithms with recombination are typical in evolutionary robotics and neuro-evolution, here we use mutation only --- following the majority of work in LONs --- but return to the question of whether crossover should be used later. At each iteration during the LON construction process, a local optimum is obtained; an acceptance condition then dictates whether it is accepted as the new incumbent solution, or whether the previous local optimum should be obtained. In this way, a process reminiscent of a Markov chain is followed. We gather data from 30 runs of ILS, each commenced from random starting solutions during the LON construction process.  ILS combines local search with random perturbations. The local search employs a first-improvement pivot rule. This means that whenever an improving (or equal) neighbour is found, it immediately becomes the incumbent genotype \cite{tari2021local}; this is in contrast to best improvement, where the best possible neighbour is found. This would require too much computation here. In the experiments, the local search process stops when there has been 100 operations applied with no improvement. Each of the runs terminates after 30 iterations with no improvement in local optimum quality or after 100 iterations in total. The local search conducts random mutation on both the robot controller and the robot design according to pre-defined probabilities described in the next section. During the runs, every accepted local optimum is hashed and logged as a network node, and any transition between local optima is logged as network edges. When a local optimum is added, its fitness, an index, and its hashed phenotype are noted too, in addition to the hashed genotype. Edges are recorded as the combination of the source node index and the destination node index, and weight is logged as well. Edge weights are the number of times a transition was followed. The LON is an amalgamation of the nodes and edges from the separate runs. 

\subsubsection{Mutation operators}\label{mutationoperators}

We use the mutation settings which were tuned in previous literature \cite{veenstra2020different}; these are provided in Table~\ref{tab:mut-params}. Perturbation is the mutation operation applied three times. Only improving or equal local optima are accepted during the search. As mentioned, both the controllers and the designs are mutated and each has a separate application rate (see Table~\ref{tab:mut-params}). For the design, the approach to mutation depends on the encoding type:

\paragraph{Direct encoding.} Mutation on the direct robot design encoding has three possible operations: removing a node, adding a node to the tree structure, and modifying a module's node. For each possible location for adding a node, a node is added according to the mutation rate; for removals, the mutation rate is halved (owing to the fact that sub-nodes are removed alongside the original node; that is, it is a more dramatic mutation). Finally, the module's nodes are modified by mutating the shape (width, height, radius) and connection angle using Gaussian mutation.

\paragraph{L-System.} Mutation on the L-System robot design encoding takes place in the following way: the rules are changed by adding or removing substitution symbols and the shape (width, height, radius) and angle of the modules in the list are mutated using Gaussian mutation. 

\paragraph{CPPN} For the CPPN encoding, the mutation is the default NEAT mutation from the \textsc{neat-python}\footnote{\url{https://neat-python.readthedocs.io/en/latest/}} library. This has several operations modifying the network structure: adding a connection; deleting a connection; adding a node; deleting a node; and replacing bias. All of these are carried out according to the robot design mutation rate. There are a large number of parameters associated with the CPPN encoding. We use the values provided in a configuration file\footnote{\url{https://github.com/FrankVeenstra/gym_rem2D/blob/master/ModularER_2D/NeuralNetwork/config}} and associated with previous literature \cite{veenstra2022effects}. As well as the CPPN mutation, the shape (width, height, radius) and angle of the modules in the list are mutated according to the mutation probability. 

\paragraph{Controller mutation.}
For all three encodings, the same controller representation is used (recall Equation~\ref{eq:controller}). For the direct and L-System encodings, controllers' parameters are mutated according to a mutation rate (controller rate in Table~\ref{tab:mut-params}) using Gaussian mutation. The standard deviation of the Gaussian distribution is set at 0.2 (the same as in~\cite{veenstra2020different}). For the CPPN, the process is different: in that case, each of the four parameters which define a controller is an output from the network (as described in Section \ref{encodings-description} and with Figure \ref{fig:cppn-dia}); they are therefore affected when the NEAT mutation is carried out on the CPPN. 

\begin{table}[h]
\centering
\caption{Mutation rates in the LON construction algorithm}\label{tab:mut-params}
\resizebox{0.28\textwidth}{!}{\begin{tabular}{lccc}
\toprule
parameter & direct & L-System & CPPN \\
\midrule
rate (controller) & 0.32 & 0.16 & 0.02 \\
rate (design) & 0.16 & 0.04 & 0.02 \\
\end{tabular}}
\end{table}

\subsection{Experimental Setup}
The majority of our experimental setup matches that of Veenstra \emph{et. al}, who compared morphological encodings~\cite{veenstra2020different,veenstra2022effects}. We begin from the associated online repository\footnote{\url{https://github.com/FrankVeenstra/gym_rem2D}} as a base for our implementation; from now on, we will refer to this as \emph{veenstra-repository}. All parameters associated with the encodings and environment are from there. For each of the three encodings, we conducted 30 independent ILS runs to log local optima and their edges; these were amalgamated to form a single LON object for each encoding. 

Statistics are calculated using the Mann-Whitney u-test \cite{mann1947test} to reject the hypothesis that two distributions are statistically the same where * means $p<0.05$, ** means $p<0.005$, *** means $p<0.0005$ and **** means $p<0.00005$ and $p$ represent the probability. 

Analysis of the LONs includes considering their weakly connected components. In directed networks, a \emph{weakly connected component} (WCC) is a group of nodes which are all reachable from each other, even if edges are uni-directional or in different directions; WCCs are identified in the analysis using Tarjan's algorithm \cite{tarjan1974new} and are referred to as \emph{components} from now on. 

\section{Results}

\subsection{Visualisation}

\begin{figure*}[tbh!]
        \centering
        \begin{subfigure}[b]{0.33\textwidth}
            \centering
            \includegraphics[trim=70 70 33 70,clip,width=\textwidth]{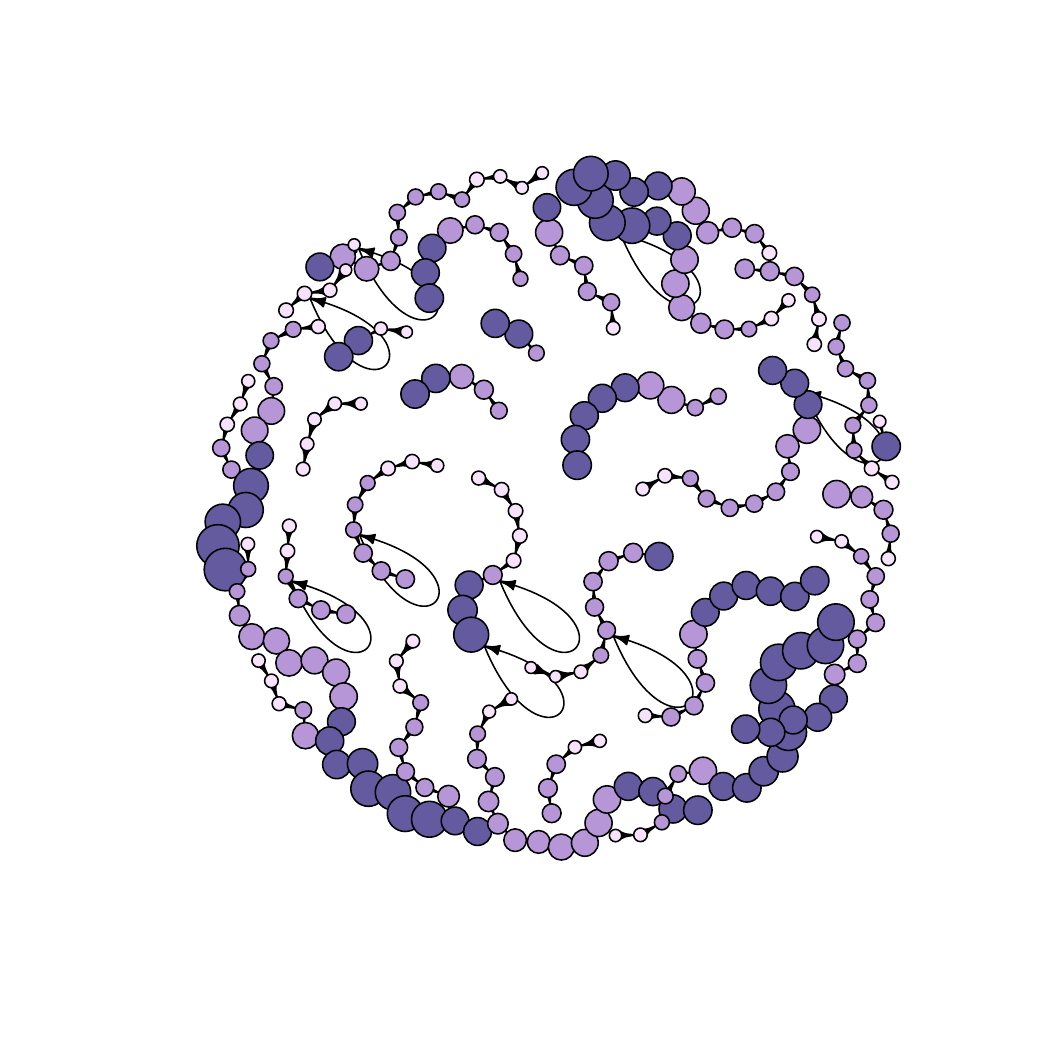}
            \caption{Direct encoding with self-loops}
            \label{fig:direct-viz}
        \end{subfigure}
        \begin{subfigure}[b]{0.33\textwidth}
            \centering
            \includegraphics[trim=70 70 70 70,clip,width=\textwidth]{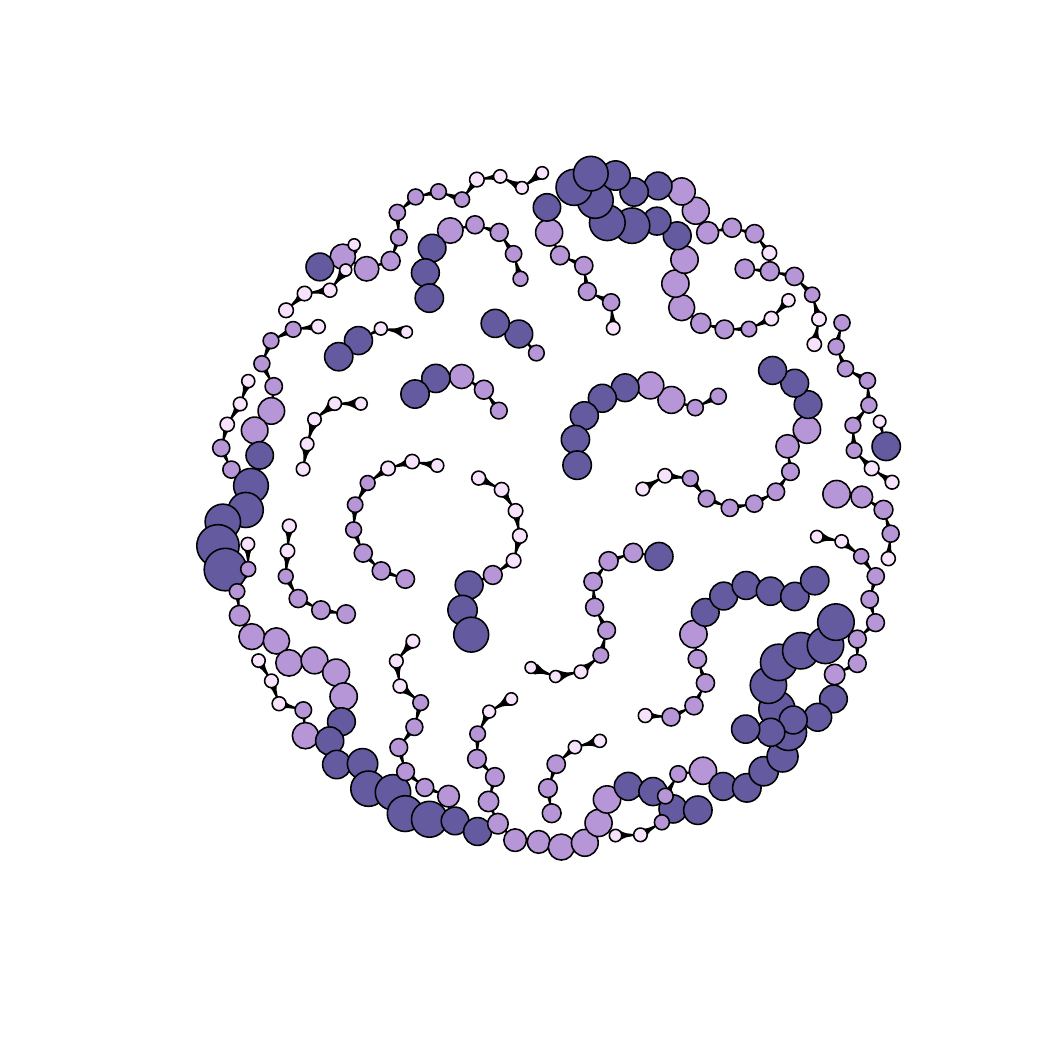}
            \caption{Direct encoding without self-loops}
            \label{fig:direct-viz-noloops}
        \end{subfigure}\\
        \begin{subfigure}[b]{0.33\textwidth}  
            \centering 
            \includegraphics[trim=70 70 70 70,clip,width=\textwidth]{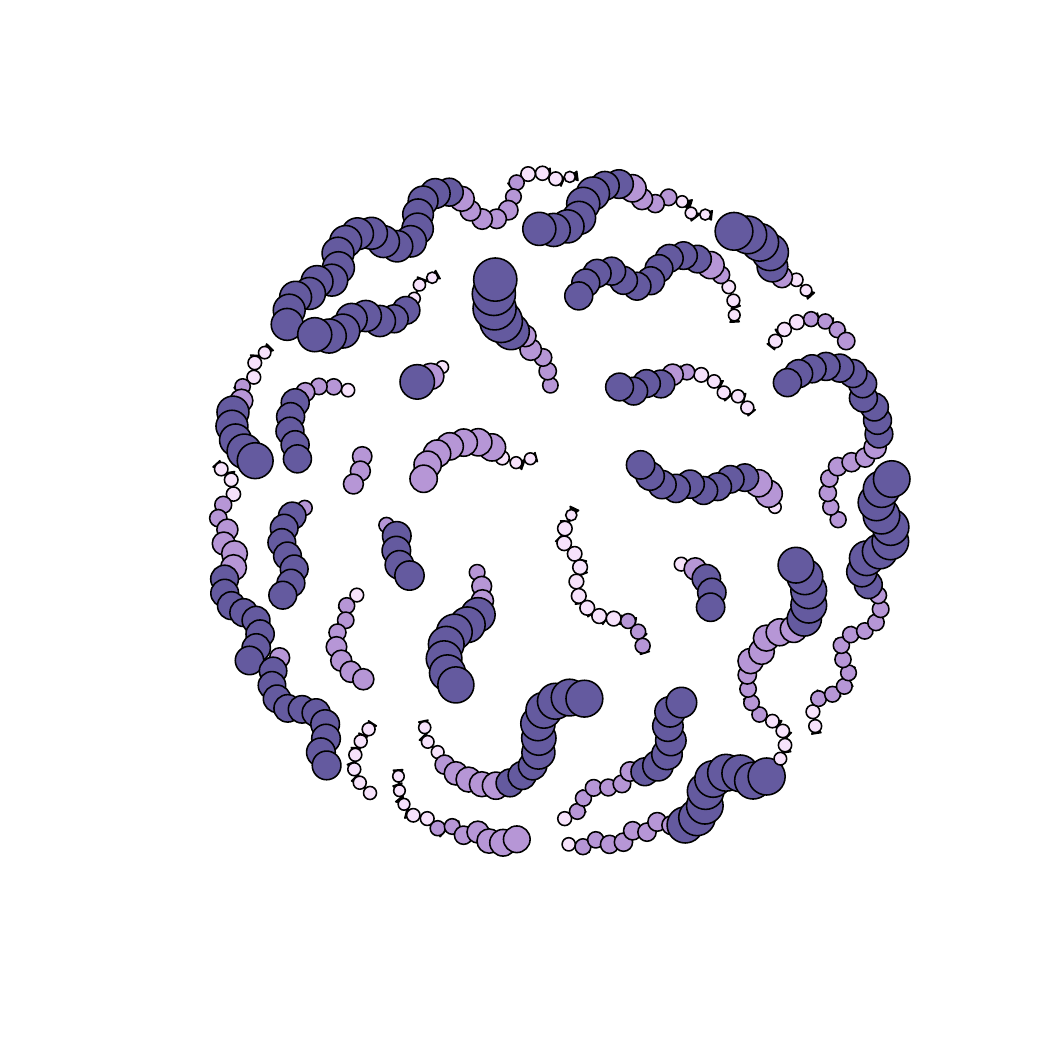}
            \caption{L-System encoding}   
            \label{fig:lsystem-viz}
        \end{subfigure}
        \begin{subfigure}[b]{0.33\textwidth}   
            \centering 
            \includegraphics[trim=70 70 70 70,clip,width=\textwidth]{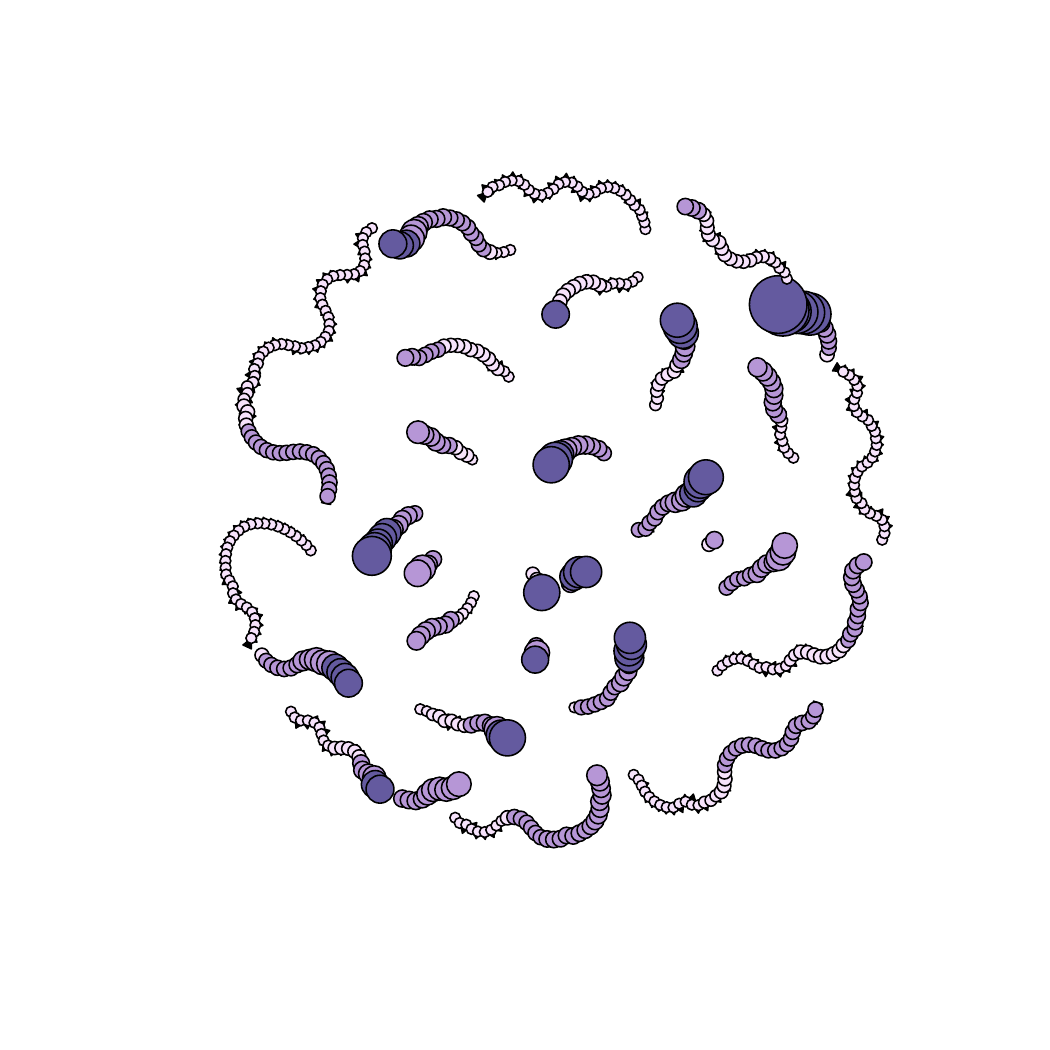}
            \caption{CPPN encoding}
            \label{fig:cppn-viz}
        \end{subfigure}
        \caption{LONs for the three robot design encodings. larger size and darker colour of nodes indicate better fitness. Node size is proportional to fitness. For colour: nodes with fitness in the first quartile of the overall fitness distribution across the three LONs (this is a fitness of less than 9.59) are low-fitness and are coloured very pale purple. Nodes within the interquartile range are middle-fitness and are light purple in colour. Finally, nodes with fitness in the upper quartile --- which is 24.48 or above --- are high-fitness and are dark purple. Self-loops are represented with edges which curve out of and back into the right of a node.}
            \label{fig:lons}
    \end{figure*}

Figures \ref{fig:direct-viz}-\ref{fig:cppn-viz} display visualisations of the extracted LONs; there is a separate Figure for each of the three robot design encoding types. In the plots, larger size and darker colour of nodes indicate better fitness. Node size is proportional to fitness. For colour, we took the approach of separating nodes into three fitness levels according to the \emph{overall distribution of fitness} across all three LONs: nodes with fitness in the first quartile (below 9.59) are low-fitness and are coloured very pale purple. Nodes within the interquartile range are middle-fitness and are light purple in colour. Finally, nodes with fitness in the upper quartile --- which is 24.48 or above --- are high-fitness and are dark purple. Self-loops are represented by edges which curve out of and back into the left of a node. The force-directed network layout algorithm \emph{kamada kawai} \cite{kamada1989algorithm} in \textsc{R} has been used to layout the LONs. Let us first consider the direct encoding LON shown in Figures \ref{fig:direct-viz} (with self-loops) and \ref{fig:direct-viz-noloops} (without self-loops). It is immediately evident from the latter that this is a network containing several isolated components; in fact, there are 30, equal to the number of ILS runs used to construct the LON. This is not a typical phenomenon seen in other domains where LON analysis has been used, e.g. \cite{treimun2020modelling,mitchell2023local,cleghorn2021understanding} which usually result in more densely connected graphs. This is likely due to the vastness of the search space. In evolutionary robotics, the search space can be essentially infinite (most controllers are encoded as real-valued numbers) and the goal is not necessarily to find a global optimum. Instead, algorithms are typically applied to find a \emph{good enough} robot that functions well. 
    
    In Figure \ref{fig:direct-viz-noloops} we can see that the 30 components are chains --- each node has either zero or one outgoing directed edges, and either zero or one incoming directed edges. We note from Figure \ref{fig:direct-viz} that the direct encoding LON exhibits several self-loops: these are recorded when an escape attempt from a local optimum is made, but returns to the same local optimum. Self-loops are not present for the LONs of the other two encodings, indicating that it may sometimes be difficult to escape local optima when using the direct encoding (i.e., there are strong attractors present). Additionally, the chains of the direct encoding (Figure \ref{fig:direct-viz-noloops}) seem to be shorter when compared to the other two encodings. This indicates a lack of \emph{navigability} associated with this encoding: it may struggle to explore genotypes (indeed, we can see from Table \ref{tab:run-stats} that this encoding resulted in far fewer unique designs being considered than the L-System). Despite this, we notice from Figure \ref{fig:direct-viz-noloops} that the direct encoding chains often terminate with a good-fitness (dark purple) local optimum. Looking across Figures \ref{fig:direct-viz}-\ref{fig:cppn-viz}, the divergence in fitness distributions can be seen. The CPPN LON in Figure \ref{fig:cppn-viz} contains a lot of nodes which are very pale purple, indicating their fitness is in the first quartile of the three-LON fitness distribution. Comparing that observation with Figure \ref{fig:lsystem-viz}, where the LON for the L-System encoding is shown, we note that there are more chains which contain the dark purple; that is, high-fitness (in the upper quartile) nodes are reached in more searches. The chains for this LON appear to be longer than those evident in the direct encoding LON (Figure \ref{fig:direct-viz}), and they very often end up with fitness in the third quartile (above 24.48). These observations, when taken together, indicate the \emph{navigability} associated with an L-System encoding. 
    
    Another observation from the CPPN LON in Figure \ref{fig:cppn-viz} is that some of the chains are lengthy but do not display any notable improvement in fitness. This tells us that the fitness landscape associated with using this encoding is not easily navigable: there appear to be large poor-quality (where poor-quality is having a fitness in the first quartile: less than 9.59) local optima plateaus which are difficult to escape (under the studied search operations and parameters). Looking at the direct and L-System encodings in Figures \ref{fig:direct-viz} and \ref{fig:lsystem-viz}, and remark that they appear to be associated with better fitness landscape navigability than the CPPN: the chains in their LONs more often display incremental improvements in fitness, which can be seen in the increase in size of node throughout a chain. This is particularly evident with the L-System LON in Figure \ref{fig:lsystem-viz}. It seems that the search becomes trapped at low-quality local optima with these encoding less often. This helps to explain the higher performance which has been observed using L-System design encoding in the literature \cite{veenstra2022effects}.

\subsection{Features}
In this Section we will first consider statistics about the ILS runs conducted with the different encodings. Table \ref{tab:run-stats} shows the rate of mutation acceptance; the rate of design acceptance (this is computed as the number of accepted mutations which were associated with a changed design); the number of unique designs present; and the total number of attempted mutations. From Table~\ref{tab:run-stats}, we notice that the mutation acceptance is lowest for direct encoding and the CPPN encoding has the lowest design acceptance. The highest rates are associated with the L-System encoding. The other two encodings tried out substantially lower numbers of unique designs than the L-System, despite having a larger number of attempted mutations. These facts, taken together, help to ratify the observations in the previous section that the direct encoding struggles to widely explore and find fitness improvement, while the L-System encoding facilitates wide exploration and moving towards better fitness. 

\begin{table}[h]
\centering
\caption{ILS run statistics for the three encodings}\label{tab:run-stats}
\resizebox{0.30\textwidth}{!}{\begin{tabular}{lccc}
\toprule
metric & direct & L-System & CPPN \\
\midrule
\emph{mutation acceptance} & 10.52\% & 47.65\% &  33.77\% \\ 
\emph{design acceptance} & 32.15\% & 38.10\% & 19.38\%  \\ 
\emph{unique designs} & \num{165800} & \num{227411} & \num{147443} \\ 
\emph{attempted mutations} & \num{352438} & \num{297964} & \num{328970} \\
\end{tabular}}
\end{table}

Table \ref{tab:lon-stats} reports measurements for the LONs associated with each encoding type. We introduce now the metrics whose names or meanings may not be intuitive: \emph{components} is the number of weakly connected components in the network. The metric \emph{path length} is the average edge distance between any two nodes in the network (excluding nodes which are unreachable to each other); \emph{degree} is the average degree of a node, including self-loops; and \emph{infeasible} is the percentage of nodes which had fitness indicating that the robot activated the killswitch and can therefore be considered infeasible. 

\begin{table}[h]
\centering
\caption{LON statistics}\label{tab:lon-stats}
\resizebox{0.25\textwidth}{!}{\begin{tabular}{lccc}
\toprule
metric & direct & L-System & CPPN \\
\midrule
\emph{nodes} & 269 & 360 & 546 \\ 
\emph{edges} & 248 & 330 & 516 \\ 
\emph{components} & 30 & 30 & 30 \\
\emph{path length} & 4.37 & 6.32 & 12.66 \\
\emph{degree} & 1.84 & 1.83 & 1.89 \\
\emph{infeasible} & 0\% & 0.003\% & 34.43\% \\
\end{tabular}}
\end{table}

\begin{figure*}[tbh!]
        \centering
         \begin{subfigure}[b]{0.31\textwidth}
            \centering
            \includegraphics[width=\textwidth]{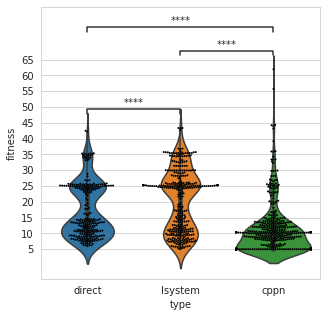}
            \caption{All fitnesses in the LON}
            \label{fig:all-fit-dist}
        \end{subfigure}
        \begin{subfigure}[b]{0.31\textwidth}
            \centering
            \includegraphics[width=\textwidth]{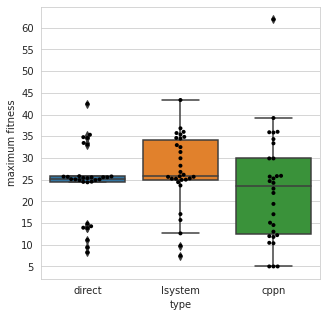}
            \caption{Maximum fitness per run}
            \label{fig:max-fit-dist}
        \end{subfigure}
        \begin{subfigure}[b]{0.31\textwidth}  
            \centering 
            \includegraphics[width=\textwidth]{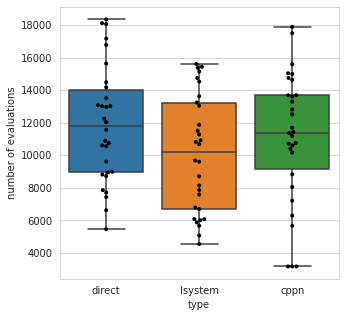}
            \caption{Number of evaluations}   
            \label{fig:evaluations}
        \end{subfigure}
        \begin{subfigure}[b]{0.31\textwidth}   
            \centering 
            \includegraphics[width=\textwidth]{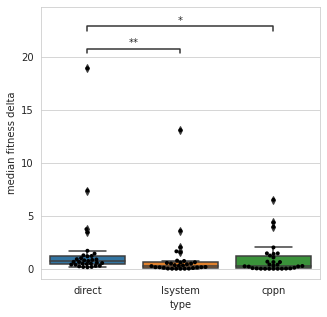}
            \caption{Fitness delta}
            \label{fig:deltas}
        \end{subfigure}
            \begin{subfigure}[b]{0.31\textwidth}   
            \centering 
            \includegraphics[width=\textwidth]{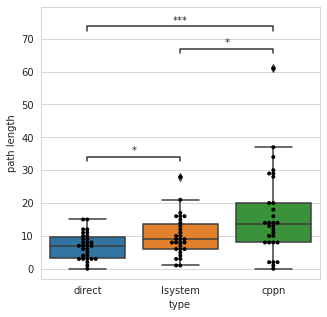}
            \caption{Chain length}
            \label{fig:chain-lengths}
        \end{subfigure}
        \caption{Distribution for various run and LON metrics across the three three encoding types. If there is a significant difference (according to a u-test) between two distributions, this is indicated with a horizontal bar and $p$-value level: *$p<0.05$; **$p<0.005$; ***$p<0.0005$; ****$p<0.00005$}
            \label{fig:distributions}
    \end{figure*}

The plots in Figure \ref{fig:distributions} show distributions for run and LON metrics across the three three encoding types. For each, there is a boxplot with a horizontal line indicating the median. We also overlay as a swarm the actual data points. For each pair of distributions, the u-test result is shown. Figure \ref{fig:all-fit-dist} displays the distribution of all fitnesses in the sampled LON nodes (local optima) of each of the three robot design encoding choices: direct, L-System, and CPPN. 

We can notice from visual inspection that the three distributions have very different medians, and the swarms show that the overall shape of the distributions are divergent to one another as well. In terms of IQR and median, the L-System has the best fitness range, with the direct encoding being next-best. This was also shown in a previous study \cite{veenstra2022effects}. The CPPN encoding is especially distinct from the other two; when including outliers, it has a much wider range, and includes more lower-quality fitnesses than those of the other two. Another interesting phenomenon is that in the case of L-System and CPPN, there appear to be groups of local optima with the same fitness: at approximately 25 for the L-System and direct encodings, and approximately 10 for the CPPN. Despite having the lowest IQR (a finding which ratifies those in the literature \cite{veenstra2020different}), the CPPN encoding resulted in the highest fitness sampled. Figure \ref{fig:max-fit-dist} shows distributions for the \emph{maximum} fitness found in each run. We can see from visual inspection that the CPPN is associated with the widest range. Indeed, it has both the highest value and the lowest. The encoding which shows the highest (best) IQR and median is the L-System type. The plot in Figure \ref{fig:evaluations} shows, for each of the three encodings, the distribution for the number of fitness evaluations used by the 30 runs. Notice from comparing the boxes visually that the distributions are similar for all three encodings. 

In Figure \ref{fig:deltas} we present the data concerning fitness deltas. A delta is here defined as the fitness increase associated with a directed LON edge (recall that the edges are always directed towards equal or better fitness, so the delta is always zero or above). For the plot, we compute the \emph{median} delta within each component in the LON; these correspond to the 30 runs. Those medians are the data shown in the Figure. Observing the three encodings, we note that the direct encoding is associated with the largest fitness deltas. This can be seen with its higher IQR and median line. Additionally, the largest value in across all encodings is achieved by the direct encoding. The L-System distribution has the lowest and narrowest IQR, implying that the fitness deltas are the smallest (but perhaps most consistent in magnitude) of the three. In terms of significant difference, the L-System and CPPN pair are statistically the same; however, the direct/L-System and direct-CPPN pairs differ, with $p<0.005$ and $p<0.05$ respectively. Finally, the distributions of chain lengths in the LON (there is one for each of the 30 runs) are captured in Figure \ref{fig:chain-lengths}. A chain is the path of local optima followed by one run of the ILS. We observe from the plot that the direct encodings had the shortest chains; this was followed by the L-System, and then CPPN with the longest chains.  
The direct-L-system pair has an associated $p<0.05$, the L-system-CPPN pair are $p<0.05$ different and the direct-CPPN are statistically different with $p<0.0005$.

\subsection{The Elephants in the Room}
One consideration of our approach is that we did not use a crossover operator. The reader may wonder whether the ILS used here actually explores promising regions of the search space, or whether instead all the sampled local optima are poor when compared to those that might be found by crossover-based approaches. To this end, we note that the fitness levels reached as shown in Figure \ref{fig:max-fit-dist} are similar to those shown in previous literature --- where an evolutionary algorithm with crossover was used \cite{veenstra2022effects} with the same virtual environment and both direct and L-System encoding. It is therefore reasonable to assume that the ILS approach reaches fitness levels approximately equivalent to those achieved by a crossover-based algorithm, and that at least some of the local optima sampled are of acceptable quality. Another factor to consider when interpreting the results is that we used mutation parameters from the literature (\cite{veenstra2020different}). Those were tuned for an algorithm which also included crossover, so it is probable that they are not the optimal settings when applied in an iterated local search. This being said, being consistent with the parameters from previous work is a reasonable place to start when beginning a new line of research. In this study, we looked at different possible genotype encoding to explore one phenotype space: a tree-based structure. Therefore, the present findings are limited to this phenotype space. In future work, it will be interesting to look at how different encodings explore different design representations such as voxel-based.

\section{Conclusions}
We have conducted a study with the aim of providing insight into performance differences associated with different robot design encodings which have been observed in the literature. To this end, we carried out fitness landscape analysis with local optima networks (LONs). Three design encodings for a virtual 2D robot were considered: a direct tree structure representation, an L-System, and a type of neural network (a CPPN). We sampled the LONs associated with each of these, visualising them and comparing their features. The results showed that an L-System encoding is associated with good fitness landscape navigability: search has the ability to escape local optima and discover a large number of unique designs. In contrast, although being able to produce a large number of unique designs, the CPPN encoding appears to be linked with low-quality local optima from which it is hard to escape towards better fitness. This observation indicates that CPPN encoding is not suited for creating the structure used to define a robot morphology in this study: a tree-based structure. We also found that a direct encoding can lead to search struggling to leave local optima; however, good fitnesses can sometimes be found when it does. These findings provide insight into previously-observed performance differences between encodings in the literature. We therefore propose that LONs can be a valuable tool for explaining phenomena seen in observed in evolutionary robotics search can inform the result of better algorithms and operators in future. The code and data associated with this work will be published in an online Zenodo repository upon acceptance. 

\bibliographystyle{ACM-Reference-Format}
\bibliography{morph}

\end{document}